%% file: acl_latex.tex
\pdfoutput=1

\documentclass[11pt]{article}

\usepackage[]{acl}

\usepackage{times}
\usepackage{latexsym}

\usepackage[T1]{fontenc}

\usepackage[utf8]{inputenc}

\usepackage{microtype}

\usepackage{inconsolata}

\usepackage{graphicx}

\usepackage{algorithm}
\usepackage{algorithmic}
\usepackage{booktabs}
\usepackage{amsmath}
\usepackage{amsfonts}
\usepackage{xspace}
\usepackage{enumitem}
\newcommand{\method}{\texttt{GraphReason}\xspace}
\newtheorem{problem}{Problem}

\title{GraphReason: Enhancing Reasoning Capabilities of Large Language Models through A Graph-Based Verification Approach}

\author{Lang Cao \\
  University of Illinois Urbana-Champaign \\
  Department of Computer Science \\
  \texttt{langcao2@illinois.edu} \\}

\begin{document}
\maketitle

\begin{abstract}
Large Language Models (LLMs) have showcased impressive reasoning capabilities, particularly when guided by specifically designed prompts in complex reasoning tasks such as math word problems. These models typically solve tasks using a chain-of-thought approach, which not only bolsters their reasoning abilities but also provides valuable insights into their problem-solving process. However, there is still significant room for enhancing the reasoning abilities of LLMs. Some studies suggest that the integration of an LLM output verifier can boost reasoning accuracy without necessitating additional model training. In this paper, we follow these studies and introduce a novel graph-based method to further augment the reasoning capabilities of LLMs. We posit that multiple solutions to a reasoning task, generated by an LLM, can be represented as a reasoning graph due to the logical connections between intermediate steps from different reasoning paths. Therefore, we propose the Reasoning Graph Verifier (GraphReason) to analyze and verify the solutions generated by LLMs. By evaluating these graphs, models can yield more accurate and reliable results.Our experimental results show that our graph-based verification method not only significantly enhances the reasoning abilities of LLMs but also outperforms existing verifier methods in terms of improving these models' reasoning performance.
\end{abstract}

\section{Introduction}

\begin{figure}[t!]
\centerline{
\resizebox{.47\textwidth}{!}{
\includegraphics{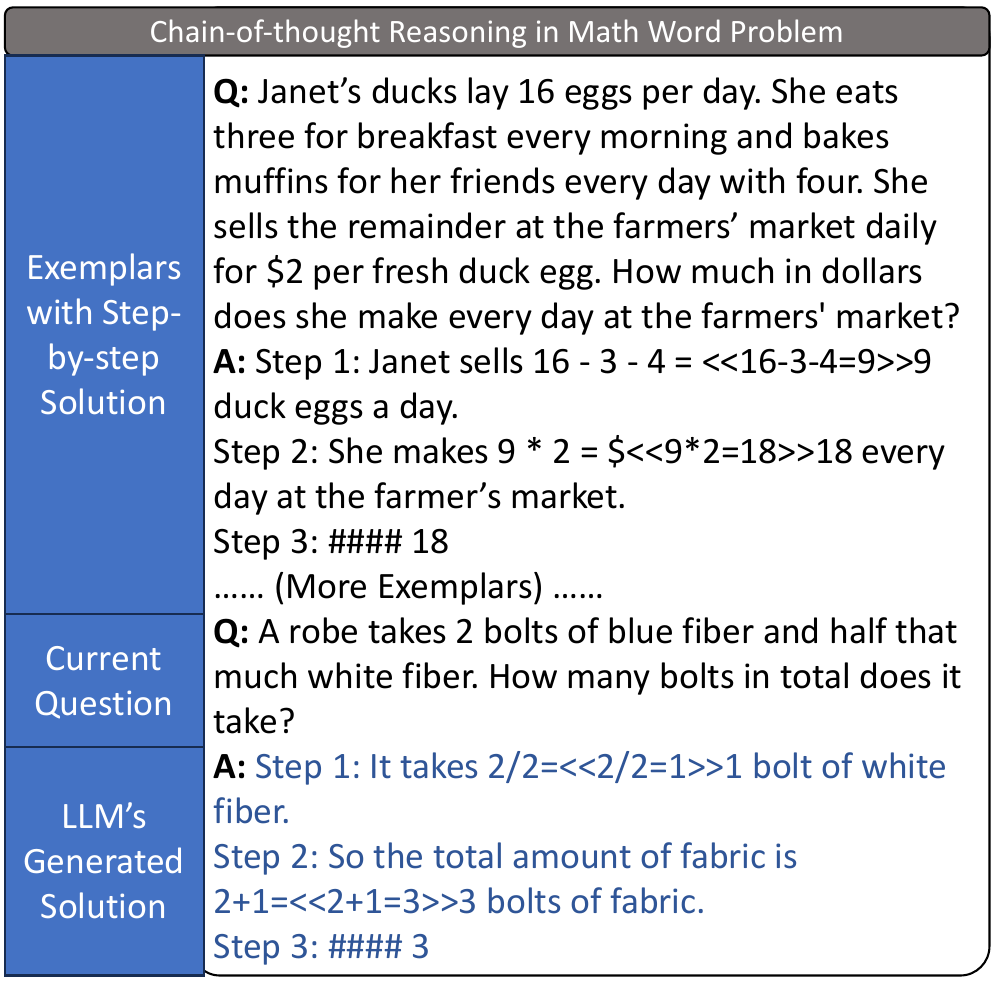}
}
}
\caption{An example of chain-of-thought reasoning in a math word problem, using data from the GSM8K dataset. Large language models learn from exemplars that provide step-by-step solutions, subsequently generating their reasoning path for the current question.}
\label{fig:cot}
\end{figure}

Large Language Models (LLMs) have demonstrated exceptional capabilities in a variety of human tasks \cite{zhao2023survey}. Among the many abilities LLMs possess, their reasoning capacity is of paramount importance \cite{kojima2023large,huang2023reasoning}. This has been substantiated by recent progresses \cite{wei2022chain,zhou2023leasttomost,lampinen-etal-2022-language}. Equipped with the ability to reason, especially in a multi-step manner, LLMs can decompose complex problems into simpler tasks, thereby facilitating their resolution. In everyday life, many complex tasks typically require multi-step solutions. A prime example of a reasoning task is arithmetic reasoning, also known as solving math word problems \cite{zhang2019gap}. These math word problems represent simplified versions of complex real-life situations.

The reasoning ability is inherent in Large Language Models (LLMs), but it necessitates specific methods for manifestation. To activate the robust reasoning capability of LLMs, the use of specially designed prompts should be considered. Numerous methods have been proposed to tap into this potential, among which chain-of-thought reasoning \cite{wei2022chain} and in-context learning \cite{lampinen2022can} are two notable approaches. Chain-of-thought reasoning can elucidate the reasoning paths during the process. In-context learning furnishes LLMs with exemplary cases, thereby enabling them to learn from and simulate these examples for improved results. In the arithmetic reasoning scenario, GPT-4 can achieve an accuracy of 92\% on the GSM8K dataset using 5-shot chain-of-thought prompts \cite{cobbe2021gsm8k}. This represents a level of difficulty that a bright middle school student should be capable of handling. As depicted in Figure~\ref{fig:cot}, this illustrates a multi-step arithmetic reasoning process in LLMs.

In addition to further training of LLMs and prompt design, some methods have been proposed to enhance the reasoning capabilities of LLMs from the perspective of output verification. The primary idea is to have LLMs generate reasoning paths multiple times, and then design a verifier to evaluate these paths and deliver the final results. \cite{wang2023selfconsistency} introduces the concept of \textit{self-consistency}, based on the intuition that a complex reasoning problem usually allows for multiple thought processes, all leading to a unique correct answer. \cite{li-etal-2023-making} also proposes \textit{All Roads Lead to Rome}, which introduces a step-aware verifier to analyze reasoning paths not just through the entire path, but at every step. However, both methods treat each reasoning path as an independent entity and do not consider the potential interrelation and interaction between different reasoning paths. Once reasoning paths are disassembled into steps, intermediate steps from one path may bear reasoning relations to other reasoning paths. These methods do not perceive all LLM outputs for a given input as a collective entity, thereby failing to analyze the internal relations of all candidate paths in depth.

Inspired by these observations, we propose \textbf{Reason}ing \textbf{Graph} Verifier (\method) in this paper. We posit that reasoning paths of one question can form reasoning graphs, where similar intermediate reasoning steps can be merged into the same node. With a graph structure, we can more effectively model and capture the reasoning logic between intermediate steps from different reasoning paths. Specifically, we first construct a reasoning graph based on all outputs from LLMs, and then train a verifier to learn the relationship between the graph structure and the final answer. During the prediction stage, we process the data in the same way as in the training stage, and use the verifier to evaluate each reasoning graph. We then select the reasoning graph with the highest score, using its answer as the final answer. To the best of our knowledge, we are the first to approach reasoning logic of LLMs from a graph perspective. We conduct extensive experiments to demonstrate the improvements over the original LLMs, and show that our method outperforms other verifiers.

In summary, our contributions are as follows:
\begin{itemize}[leftmargin=*, itemsep=0pt, labelsep=5pt]
    \item We propose a graph-based verification method, \method, aimed at significantly enhancing the reasoning capabilities of large language models without the need for additional training of LLMs.
    \item We establish an arithmetic reasoning benchmark using three Math Word Problem datasets to illustrate the fundamental reasoning performance of large language models, and to provide a fair comparison of the performance of various existing verifiers.
    \item Our experimental results indicate that the method proposed in this paper outperforms other enhancement methods. We also provide an extensive analysis of the limitations and future potential of \method.
\end{itemize}

\section{Related Works}

\noindent\textbf{Reasoning of Fine-tuning Models} has been extensively studied. It focuses on addressing reasoning tasks using a general sequence-to-sequence approach, enhanced by reasoning-aware pre-training or fine-tuning of language models. \cite{cobbe2021gsm8k} proposed training a verifier to rank solutions sampled from fine-tuned language models. \cite{yoran-etal-2022-turning,wang-etal-2022-logic} suggested equipping language models with reasoning abilities by generating training examples with human-designed templates. \cite{pi-etal-2022-reasoning} proposed injecting reasoning capabilities into language models by continually pre-training on program execution data.

Several studies have focused on imbuing PLM with reasoning ability for specific tasks, such as arithmetic reasoning \cite{cobbe2021gsm8k,miao-etal-2020-diverse,patel-etal-2021-nlp}, commonsense reasoning \cite{talmor-etal-2019-commonsenseqa}, and inductive reasoning \cite{sinha-etal-2019-clutrr}. For instance, various strategies have been proposed to improve language models' performance on arithmetic reasoning tasks, often referred to as math word problems. \cite{ijcai2019p736} proposed a tree-structured decoder to generate an equation tree, while \cite{zhang-etal-2020-graph-tree} applied graph convolutional networks to extract relationships of quantities in math problems. \cite{li-etal-2022-seeking} used contrastive learning to better learn patterns in math word problems. However, \cite{valmeekam2023large,rae2022scaling} suggested that reasoning, particularly multi-step reasoning, is often a weakness in language models and other NLP models. \\

\noindent\textbf{Reasoning of Large Language Models} has garnered significant attention and demonstrated immense potential. Recent advancements in LLMs suggest that the ability for multi-step reasoning is already embedded within these large-scale models \cite{kojima2023large,huang2023reasoning}, such as PaLM \cite{chowdhery2022palm}, GPT-4 \cite{openai2023gpt4}. Therefore, providing an adequate prompt is sufficient to utilize this reasoning ability. For example, the prompting method proposed by \cite{kojima2023large,wei2022chain}, which is based on a chain-of-thought, could aid LLMs in generating text with arithmetic reasoning and common factual knowledge. Following \cite{wei2022chain}, experiments on current language models demonstrated that chain-of-thought prompting could enhance the accuracy of solving math problems from 18\% to 57\%. \cite{lampinen2022can} included explanations in the in-context examples and tested the influence of explanations by evaluating the score between \textit{explain-then-predict} and \textit{predict-then-explain}. Moreover, \cite{zhou2023leasttomost} suggested a two-stage prompting strategy, \textit{least-to-most} prompting, which breaks down a complex problem into a series of subproblems and solves them step-by-step. \cite{li-etal-2023-making} proposed sampling multiple times from diverse prompts to enhance the variety of responses.

\begin{figure*}[t!]
\centerline{
\resizebox{\textwidth}{!}{
\includegraphics{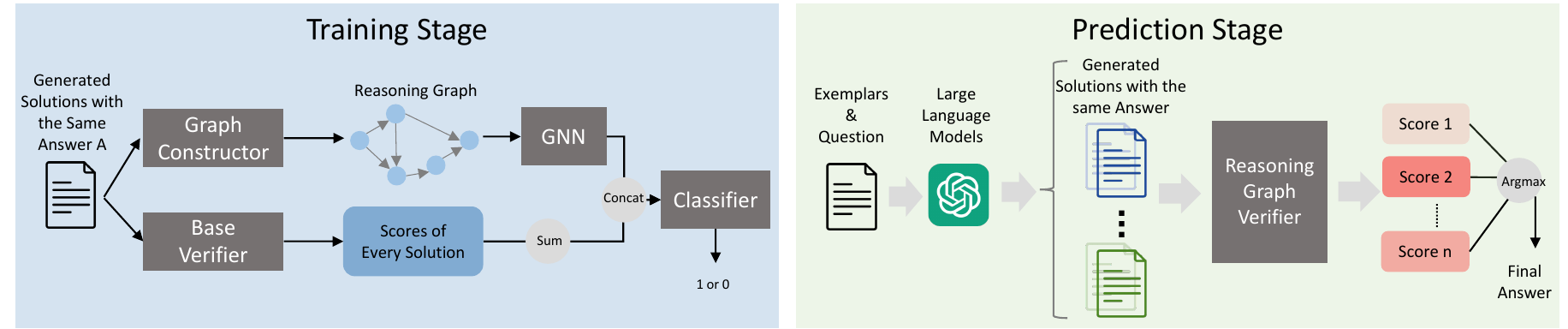}
}
}
\caption{The framework of \method. In the training stage, \method processes generated solutions from LLMs to construct reasoning graphs, and then trains a verifier to judge them according to graph classification. In the prediction stage, \method evaluates candidate solutions to assign a score, and selects the solution with the highest score as the final answer.}
\label{fig:framework}
\end{figure*}

In addition to designing prompts, adopting additional strategies like verifier has contributed to enhancing the performance of reasoning abilities of large language models. For instance, \cite{wang2023selfconsistency} proposes \textit{self-consistency}, which involves sampling different reasoning paths from the language model, and then returning the most consistent final answer via majority voting. \cite{li-etal-2023-making} used a step-aware voting verifier to enhance the reasoning ability of LLMs from two perspectives. These methods strive to augment the reasoning abilities or yield superior reasoning results without necessitating additional training of LLMs. Our work continues this research direction, with a specific focus on developing a novel graph-based verifier to boost the reasoning capabilities of LLMs.

\section{Methodology}
\subsection{\method Framework}
\begin{problem}[Reasoning to Solve Problems]
Given a set of $n$ math word problems $\mathbf{Q} = \{Q_1, Q_2, ..., Q_n\}$, where each $Q_i$ is represented by the text description of a single math word problem, the goal of reasoning to solve math word problems is to generate the answers $\mathbf{A} = \{A_1, A_2, ..., A_n\}$ for these problems. Here, each $A_i$ represents the generated text of the corresponding answer. During the process of large language models generating answers, a set of $n$ reasoning paths for solutions $\mathbf{S} = \{S_1, S_2, ..., S_n\}$ is also produced. Each solution $S_i$ is represented as $S_i = \{Q, Step_1, Step_2, ..., Step_l, A\}$, where each $Step_i$ denotes the intermediate steps in the step-by-step solutions.
\end{problem}

We propose \method to verify the solutions generated by LLMs in order to improve the final answer accuracy. This method is a graph-based verification technique that analyzes reasoning paths from generated solutions from a graph perspective. The final answer is obtained without modifying the original LLMs, functioning much like a plugin. As illustrated in Figure~\ref{fig:framework}, there are two steps in the training stage: \textit{Graph Construction} and \textit{Graph Classification}. In the \textit{Graph Construction} step, we obtain the generated solution from LLMs with the specific designed prompt and group them according to their final answers. We split reasoning paths by steps and then merge intermediate steps with identical expression to form reasoning graphs. In the \textit{Graph Classification} step, we classify these reasoning graphs with the additional feature of the sum of scores from the \textit{base verifier} to train the integrated verifier model. In the prediction stage, the candidate solutions are first generated by LLMs. We process them in the same manner as in the training stage, then we use trained verifier to evaluate the scores of each candidate solution. The best solution, denoted by the highest score, is selected as the final predicted answer. We will now provide a detailed introduction to the entire process.

\subsection{Prompt Design}
To improve the output of Language Models (LLMs) in providing solutions, it is essential to design effective prompts. We incorporate chain-of-thought and in-context learning to enable LLMs to generate step-by-step answers for math word problems. The language models generate output $\mathbf{y}$ based on the input $\mathbf{x}$ using the following equation:
\begin{equation}
p(\mathbf{y}|C, \mathbf{x}) = \prod_{t=1}^{|\mathbf{y}|} p_{LM}(\mathbf{y}_t|C, \mathbf{x}, \mathbf{y}<t),
\end{equation}
where, $C$ represents the input provided to the LLMs prior to the current math word problem's question. $C$ is a concatenation of $k$ exemplars, denoted as:
\begin{equation}
C = [(Q_1,S_1,A_1); (Q_2,S_2,A_2), ...; (Q_k,S_k,A_k)],
\end{equation}
where, $Q_i$ represents the question, $S_i$ represents the intermediate steps of the solution, and $A_i$ represents the answer. We set $k$ to five in this study, resulting in a prompt that consists of five question-answer pairs sampled from the training split of a math word problem dataset. Therefore, the prompt can be denoted as:
\begin{equation}
\mathbf{Prompt} = [C;Q],
\end{equation}
where $Q$ represents the question of the current math word problem.

Using a greedy decoding approach to sample one output from LLMs may not be robust. It can lead to instability and occasional errors. To address this, \cite{wang2023selfconsistency} propose the concept of \textit{self-consistency}. This approach involves sampling different reasoning paths from the language model and then selecting the most consistent final answer through majority voting. Instead of using greedy decoding to sample only once and verify, they utilize sampling decoding to sample $N_1$ times. We also follow the idea presented by \cite{li-etal-2023-making} in their work named \textit{All Roads Lead to Rome}. This approach involves generating $N_2$ diverse prompts for LLMs to produce multiple outputs. By employing multiple sampling decodes on diverse prompts, we can obtain generated solutions from different sources. Specifically, we obtain $N = N_1 \times N_2$ diverse reasoning paths for each question. In our main experiments, we set $N_1 = 10$ and $N_2 = 3$. These solutions will be further processed and verified using our designed verifier.

\begin{figure*}[t!]
\centerline{
\resizebox{0.9\textwidth}{!}{
\includegraphics{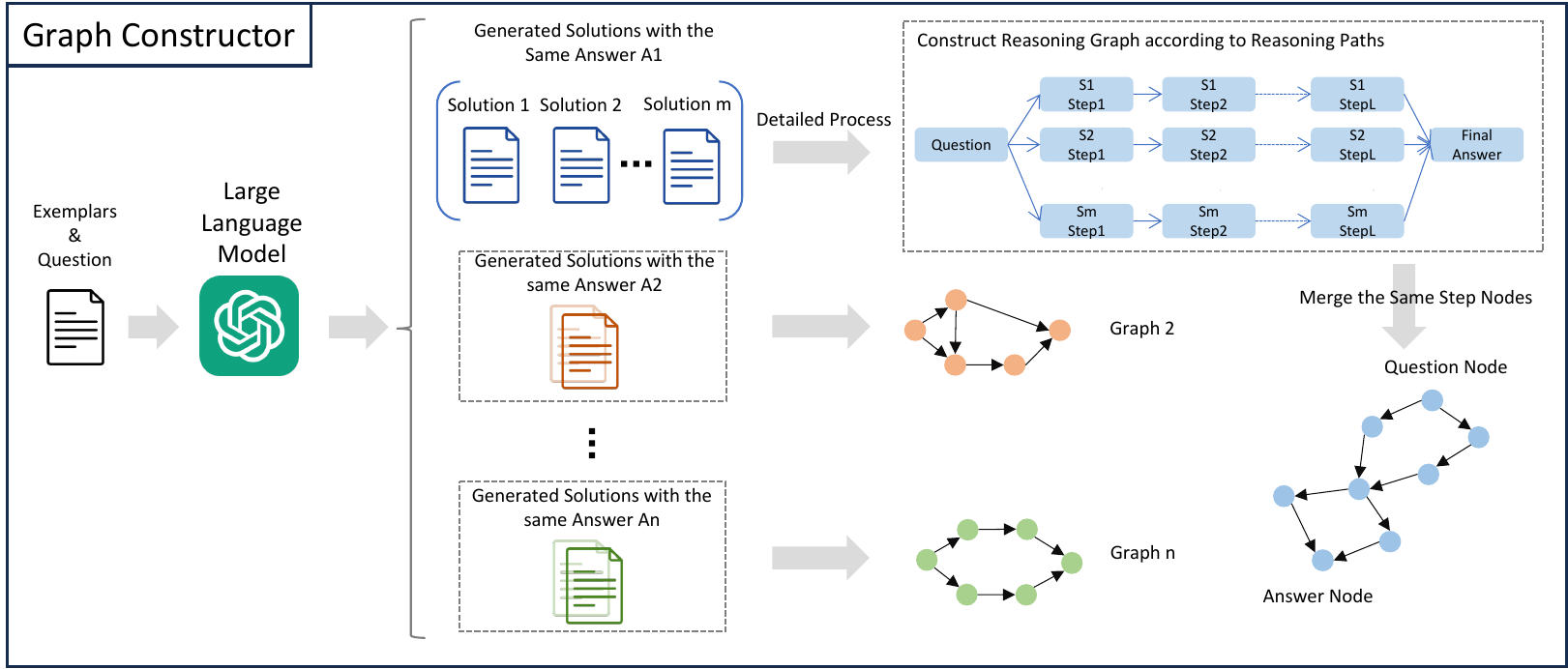}
}
}
\caption{The graph constructor in \method. We detail the process of transforming `Generated Solutions with the
Same Answer A1' to `Graph 1'.}
\label{fig:graph-constructor}
\end{figure*}

\subsection{Reasoning Graph Construction}

After generating multiple solutions for a question, it becomes necessary to construct reasoning graphs based on the reasoning paths taken by these solutions.

As shown in Figure~\ref{fig:graph-constructor}, we begin by grouping all the generated solutions for a particular question according to their final answer. Since these solutions originate from the same question, their reasoning paths will share the same starting point. Similarly, solutions with the same final answer will have the same endpoint, as their reasoning paths converge. Therefore, a group of generated solutions with the same final answer can form a reasoning graph with a uniform start node (question node) and end node (answer node). We define this division process as follows:
\begin{equation}
\begin{aligned}
\mathbf{S} = \{S_{A_1}, S_{A_2}, ..., S_{A_n}\},
\end{aligned}
\end{equation}

where $\mathbf{S}$ represents the set of generated solutions for a question, and $S_{A_i} = \{S_1, S_2, ..., S_m\}$ is the subset of generated solutions that all have the same final answer $A_i$.

\begin{figure}[t!]
\centerline{
\resizebox{.5\textwidth}{!}{
\includegraphics{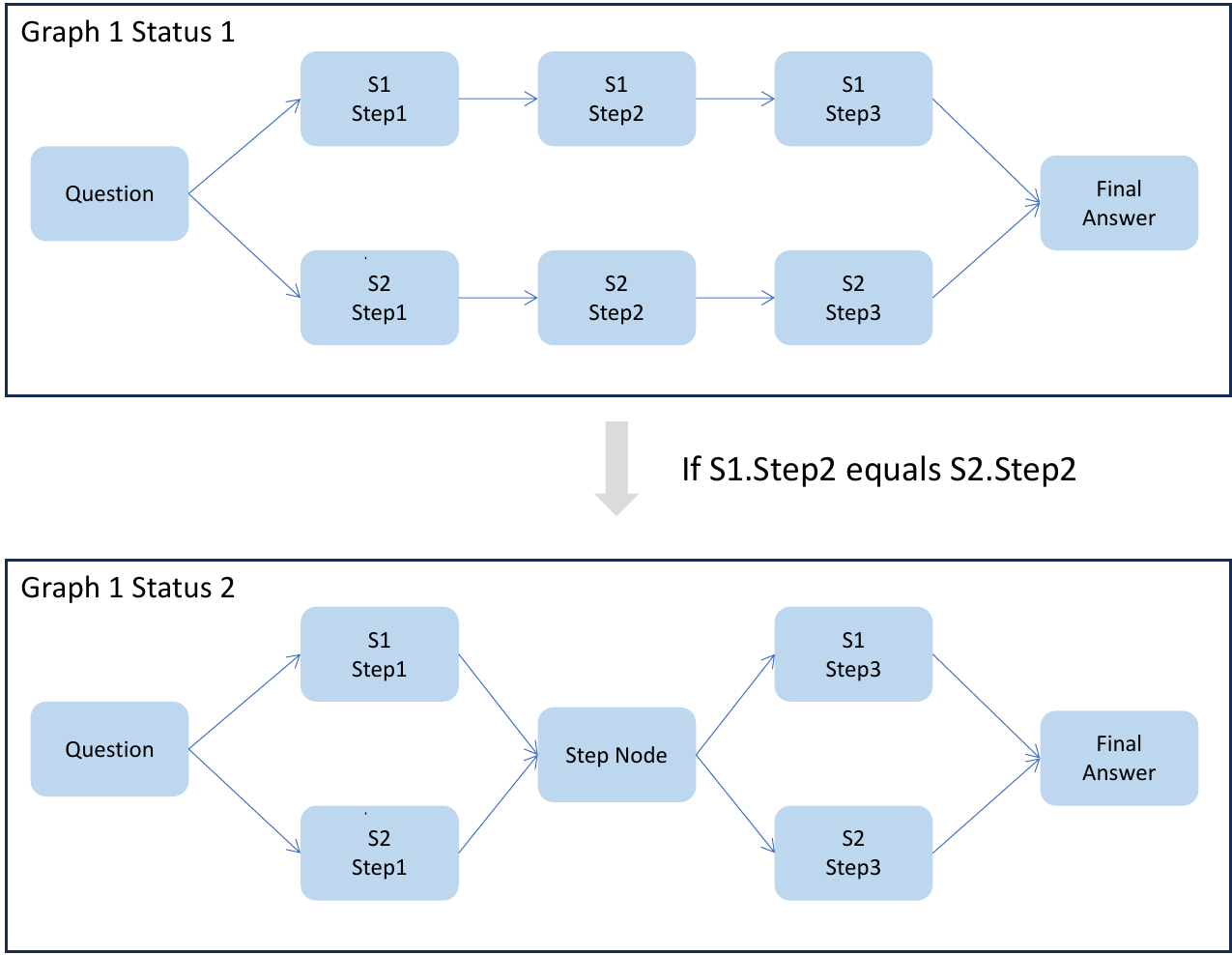}
}
}
\caption{The process of reasoning graph construction. The primary operation here is the merging of identical intermediate steps in reasoning paths into a single graph node.}
\label{fig:construct}
\end{figure}

For each subset of generated solutions $S_{A_i}$, we construct a reasoning graph. This construction is motivated by the understanding that each step in the reasoning path of a generated solution does not exist in isolation from the other solutions. The steps from one solution's reasoning path can impact the steps from another solution, enhancing the overall reasoning process. We utilize the graph structure to model and capture these relationships between steps from different solutions. As the different reasoning paths can benefit each other, we construct a reasoning graph to link these paths together. As shown in Figure~\ref{fig:construct}, the primary operation here is the merging of identical intermediate nodes in reasoning paths into a single graph node. We first compare the reasoning steps from any two solution reasoning paths. If they have the same intermediate steps of arithmetic expression, we merge them into the same node, and if they differ, we do not. For reasoning math word problems here, we define reasoning steps as the current arithmetic expression without other language text in the current reasoning step for clarity. It can help us simplify construction of reasoning graphs in the reasoning task. The detailed algorithm for constructing a reasoning graph is shown in Algorithm~\ref{alg:graph}.

\input{Algorithms/construct}

The generated solutions, divided by their final answers $\{S_{A_1}, S_{A_2}, ..., S_{A_n}\}$, can be transformed into $n$ reasoning graphs of generated solutions $\{G_{A_1}, G_{A_2}, ..., G_{A_n}\}$.

Regarding the node features in the graph, we select the score from the \textit{Base Verifier} and the node degree. We believe the score from the \textit{Base Verifier} encapsulates the semantic information of solutions, and the node degree contains information about the graph structure. The \textit{Base Verifier} is trained independently from the whole framework. It is designed to judge whether a single reasoning path of one solution is correct, which is a binary text classification task. After training, it can be used to verify any single solution and assign a $score \in (0, 1)$ to evaluate the likelihood of the solution being correct, where $score = 0.99$ suggests a $99\%$ probability of the solution being correct. We use the score from the \textit{Base Verifier} to better incorporate solution semantic information because, according to our experiments, it is challenging to model semantic information while modeling reasoning logic information. The score of a step is the same as its solution score. Therefore, for one step node $V$, it has many scores $\{score^a, score^b, ..., score^c\}$ from different solutions. The feature of one node $V_i$ in the graph is then concatenated by the selected feature, which can be represented as:
\begin{equation}
\begin{aligned}
\mathbf{V} = [score_i^{mean}, score_i^{max}, score_i^{min}, \\
score_i^{num}, in\_degree_i],
\end{aligned}
\end{equation}
where $\mathbf{V} \in \mathbb{R}^{5}$, $score_i^{mean}$ is the mean of all scores of one step $V_i$, $score_i^{max}$ is the maximum score, $score_i^{min}$ is the minimum score, $score_i^{num}$ is the number of scores, and $in\_degree_i$ is the in-degree of the step node $V_i$.

In this way, we can obtain multiple reasoning graphs to represent all generated solutions from LLMs for a single math word problem question.

\subsection{Verifier Design}
Our designed verifier \method,is used to evaluate the answer of a generated solutions group, which is also represented as a reasoning graph. This verifier has two inputs: the graph and the sum of solution scores. We employ the Graph Isomorphism Network (GIN) \cite{xu2018how} to perform node feature propagation, thereby encoding the information from the reasoning graphs we obtained. The node feature is propagated and aggregated as follows:
\begin{equation}\small
	h_v^{(k)} = MLP^{(k)}\big( (1 + \varepsilon^{(k)} ) \cdot h_v^{(k-1)}  +  \sum\nolimits_{u \in N(v)} {h_u^{(k-1)}}   \big),
\end{equation}
where $h_v^{(k)}$ represents the state of node $v$ after the $k^{th}$ update. $MLP^{(k)}$ refers to a multi-layer perceptron in the $k^{th}$ layer. $N(v)$ denotes all the neighbors of node $v$ and $\varepsilon$ is a learnable parameter.
Then, we perform a sum readout to obtain the representation of the reasoning graph:
\begin{equation}\small
	h_{G} = \sum_{v \in G} h_v^{(k)},
\end{equation}
where $h_{G}\in \mathbb{R}^{5}$. We set $k$ to 3, signifying the application of three layers of GIN. Concurrently, the sum of the scores of solutions with the same final answer, $A_i$, denoted as $score_{A_i}$, is represented as follows:
\begin{equation}
score_{A} = \sum_{i \in S_{A}} score_i.
\end{equation}
Then a reasoning graph can then be represented as:
\begin{equation}
\mathbf{G} = [h_{G}, score_{A}],
\end{equation}
where $\mathbf{G} \in \mathbb{R}^{6}$.

The target label of the graph $\mathbf{y} \in \{0, 1\}$ indicates whether the final answer matches the correct final answer. We compute the loss and train the verifier model by:
\begin{equation}
\mathcal{L} = \sum_{i = 1}^n \mathcal{L}_{BCE} (label_i, f(\mathbf{G}_i)),
\end{equation}
where $i$ represents the number of solution subset among all $n$ subsets after grouping solutions. The corresponding reasoning graph for this subset is denoted by $\mathbf{G}_i$, and $f()$ is a linear classifier.

\subsection{Answer Verification}
During the prediction stage, all generated solutions are processed in the same way as in the training stage. The trained verifier is then used to evaluate the scores of each reasoning graph, each of which represents a group of solutions that yield the same final answer. The final answer associated with the highest score is selected as our final predicted answer:
\begin{equation}
\hat{\mathbf{y}} = \mathbf{Answer}[\underset{i}{\arg\max} score_i],
\end{equation}
where $score_i$ denotes the score of the reasoning graph $\mathbf{G}_i$, as determined by our verifier. $\mathbf{Answer}$ represents the list of all candidate final answers. By predicting the number of the optimal reasoning graph, we can determine the final predicted result of the current reasoning task.

\section{Experiments}
\input{Tables/exp}

In this section, we conducted extensive experiments to demonstrate the performance of \method, along with a more in-depth analysis. Universally, we reproduced all types of verifiers to report their results based on the same generated solutions. Our experiments are conducted in two settings: Arithmetic Reasoning and Commonsense Reasoning. We ensured a fair comparison by setting the same random seed, using the same hardware environment, and applying similar hyperparameters. We used accuracy as the metric to evaluate the ability of solving math word problems, which determines whether the final answer is correct or not.

\subsection{Training Details}
For LLMs sampling, we use \textit{gpt-3.5-turbo} as our base LLMs and set the temperature $t$ to 1. All verifiers use the same LLMs' output. Regarding verifier training, we fine-tune on \textit{bert-base-uncased} \cite{devlin2019bert}. We employ the AdamW optimizer \cite{loshchilov2019decoupled} to optimize the model parameters during training. We apply differential learning rates, setting the learning rate of the final linear classifier to 4e-2, while the other graph neural network layers are set to 4e-3. The activation layer between them is ReLU \cite{agarap2019deep}. The batch size in each training step is set to 2. The batch size is small because the verifier needs to verify multiple reasoning graphs for a single question.

To ensure a fair comparison between the Voting Verifier, Simple Verifier, and \method, we use the same trained base verifier for all three approaches.

The details of the datasets and baselines are provided in Appendix~\ref{sec:dataset} and Appendix~\ref{sec:baseline}, respectively.

\subsection{Main Results}
\input{Tables/ablation}

We present the main results in Table~\ref{tab:exp}. As can be seen from the table, \method significantly enhances the original \textit{gpt-3.5-turbo}'s reasoning abilities across all three datasets, for instance, improving accuracy by 13.0\% (72.7\% $\rightarrow$ 85.7\%) on GSM8K. It is also evident that our method surpasses other verifier methods with the same output from LLMs and achieves the state-of-the-art on all three datasets.

Additionally, the Step-aware Voting Verifier improves upon the Voting Verifier by recognizing that not all steps in an incorrect reasoning path are equally erroneous, and some steps may still be useful for reasoning. We believe this hypothesis is overly simplistic and cannot describe complex logical relationships among steps. According to Table~\ref{tab:exp}, it leads to some metric decline, and the same finding also observed in the original paper. Furthermore, it does not perform well in the StrategyQA task, because there are no gold reasoning paths for the training of this commonsense reasoning task. In this task, the reasoning paths are generated and pseudo, indicating a requirement for gold labels at each step of the reasoning process. However, our paper consistently improves upon the Voting Verifier by considering complex relationship between different reasoning paths through reasoning graphs. We enhance the previous method, which did not consider relations in steps between different solutions, by 0.3\% (85.4\% $\rightarrow$ 85.7\%), 0.3\% (85.1\% $\rightarrow$ 85.4\%), 0.1\% (96.9\% $\rightarrow$ 97.0\%), and 0.5\% (70.7\% $\rightarrow$ 71.2\%) across the four datasets.

Moreover, \method yields only a slight improvement in performance on ASDiv-a, and the results are nearly identical. One reason for this is that the math word problems from ASDiv-a are simpler compared to those in the other two datasets, based on our observations. In most cases, these problems do not require complex reasoning from a graph perspective to generate a satisfactory answer. It demonstrates that our method is particularly well-suited for such situations. We believe that \method can offer more substantial improvements in the more complex scenario.

\subsection{Ablation Study}
We conducted an ablation study to evaluate the impact of each component on the overall performance of our method. Table~\ref{tab:ablation} presents the results of this study, highlighting how these modules contribute to the improvement of the base model in distinct ways. It can be observed that the omission of any component leads to a decline in the final result. The solution semantics from the base verifier appear to be most crucial to \method. The current method still relies on semantic information, which is reasonable since reasoning steps from different solutions require semantic information for better reasoning. We also notice that reasoning graphs bring a slight improvement to the entire method, thereby proving effectiveness of graph structure. The improvement is not substantial because we do not model the graph structure and semantic information simultaneously, and create a training gap here. Another essential factor is the complexity of graph classification, compounded by the presence of noise and limitations in our training data.

\subsection{\method with Different LLMs}

\input{Tables/llms}

To evaluate the compatibility of \method and its effectiveness across various models, we additionally include \textit{gpt-4} \cite{openai2023gpt4} and \textit{PaLM-2} \cite{anil2023palm} in our experiments. Given our limited computing resources, we utilize the same training data previously sampled from \textit{gpt-3.5-turbo}. For testing in the GSM8K task, we select samples from 100 pieces of data from \textit{gpt-4} and \textit{PaLM-2} respectively. We conduct the sampling 10 times using three types of five exemplars, maintaining the same settings as in our previous experiments. Our method aims to enhance the original reasoning capabilities. Therefore, we do not include small-sized LMs, which typically exhibit weaker reasoning abilities.

From Table~\ref{tab:llms}, it is evident that our method enhances the original reasoning performance of both \textit{GPT-4} and \textit{PaLM-2}. However, there is a performance decline in \textit{gpt-4} when compared with the best baselines. The performance of \method is comparable to that of the voting method. We hypothesize that this is because the reasoning patterns of \textit{GPT-4} differ from those of \textit{GPT-3.5-Turbo}, and our verifier is trained specifically on \textit{GPT-3.5-Turbo} samples in this setting.

\section{Conclusion}
In this paper, we propose \method, a novel and general method to enhance the reasoning abilities of large language models. Our method is the first to approach reasoning logic of large language models from a graph perspective and verifies candidate reasoning paths accordingly. We demonstrate the superiority of \method through extensive experiments.


\section*{Limitations}
There are several limitations in the current research that contribute to performance that is not as good as expected:
\begin{itemize}[leftmargin=*, itemsep=0pt, labelsep=5pt]
    \item \textbf{Computing Resources}. Despite the impressive performance it achieves, our framework requires large language models like GPT3.5. Inference with these models is more time-consuming and costly than fine-tuning models like BERT\cite{devlin2019bert}. Some experiments, such as hyperparameter analysis, have already been conducted in related previous work and are not replicated here. Furthermore, due to limited computing resources, we have not conducted experiments with additional LLMs. We have chosen solely to use the representative LLM, GPT3.5, to compare the performance of the verifiers.
    \item \textbf{Labeled CoT data}. \method is a complex verifier method that builds on graph classification, which requires more labeled data with well-annotated chain-of-thought reasoning paths for training. In the training of \method, we use reasoning paths from LLMs' output which may introduce significant noise. If the training data included labeled reasoning graphs, the performance would improve significantly.
    \item \textbf{Other Reasoning Tasks}. There are many types of reasoning tasks beyond math word problems, such as Commonsense Reasoning \cite{talmor-etal-2019-commonsenseqa}, Inductive Reasoning \cite{sinha-etal-2019-clutrr}, etc. Given that graph construction is a complex process, we have focused mainly on solving math word problems (Arithmetic Reasoning). This focus allows for a more convenient implementation of the merging of intermediate steps. In other cases, identifying similar steps can be challenging. On the other hand, a math word problem typically presents a greater variety of potential solutions.
\end{itemize}

Nevertheless, we believe that future studies, conducted by us or others, can overcome these limitations and further improve upon our approach.


\bibliography{custom}

\appendix

\section{Datasets}
\label{sec:dataset}
We compared \method with other methods on three different math word problem datasets: \textbf{GSM8K} \cite{cobbe2021gsm8k}, \textbf{SVAMP} \cite{patel-etal-2021-nlp}, and \textbf{ASDiv-a} \cite{miao-etal-2020-diverse} and one commonsense reasoning dataset: \textbf{StrategyQA} \cite{geva2021strategyqa}. We selected the subset ASDiv-a (arithmetic) from the original dataset ASDiv, which only involves arithmetic operations.

These three arithmetic reasoning datasets are more challenging than other math word problem datasets, making them more suitable for testing the reasoning capability of LLMs with a verifier. As the GSM8K dataset is the only one providing step-by-step solutions as chain-of-thought exemplars, we chose exemplars from the GSM8K training dataset and tested them on all three datasets. Additionally, the training data for the verifier also used the GSM8K training data. In this setting, we could also demonstrate the transfer learning and generalization ability of our method. The size of the training split from GSM8k is 1000. The test data sizes for GSM8K, SVAMP, and ASDiv-a are 1319, 1000, and 1218, respectively.

In the StrategyQA commonsense reasoning task, we set the number of exemplars to 8 and select pseudo-exemplars from \cite{li-etal-2023-making}. Additionally, we conduct five sampling iterations for each context of LLMs. From the entire dataset, we select a subset of 1,000 instances, allocating 700 for training and 300 for testing.

\section{Baselines}
\label{sec:baseline}
In our evaluation, we consider the following baselines:
\begin{itemize}[leftmargin=*, itemsep=0pt, labelsep=5pt]
    \item \textbf{Greedy Decode} is a simple method that uses a greedy decoding strategy to sample once.
    \item \textbf{Self-Consistency (Voting)} \cite{wang2023selfconsistency} samples multiple times and selects the final answers based on majority voting.
    \item \textbf{Simple Verifier} \cite{cobbe2021training}, which is also known as the Sampling and Re-ranking strategy, uses a verifier to assign scores to sampled solutions and selects the final answer with the highest score.
    \item \textbf{Voting Verifier} \cite{li-etal-2023-making} combines the Voting and Verifier approaches. It assigns total scores to answers from scores of all candidate solutions and selects the final answer with the highest score.
    \item \textbf{DIVERSE (Step-aware Voting Verifier)} \cite{li-etal-2023-making}, which is the state-of-the-art method, considers the reasoning steps throughout the entire reasoning path. It recognizes that not all steps in an incorrect reasoning path are equally wrong and that some steps may still be useful for reasoning.
\end{itemize}

We primarily compare \method with other verifiers using the same generated solutions from \textit{gpt-3.5-turbo}. Additionally, we include some previous Fine-tuning state-of-the-art methods to reflect the strong reasoning ability of LLMs. The previous Fine-tuning SOTA methods are denoted as follows: a: \cite{cobbe2021training}, b: \cite{pi-etal-2022-reasoning}, c: \cite{miao-etal-2020-diverse}, d: \cite{chowdhery2022palm}.


\end{document}

%% file: Algorithms/construct.tex
\begin{algorithm}[tb]
\caption{Reasoning graph construction algorithm}
\label{alg:graph}
\textbf{Input}: generated solutions $S_{A_i}$ which have the same final answers \\
\textbf{Output}: a reasoning graph $G_{A_i}$ \\
\begin{algorithmic}[1]
\STATE $node\_num \gets 0$
\STATE $node2id \gets dict()$
\STATE $edges \gets list()$

\FOR{each $reason\_path$ in $S_{A_i}$}
    \FOR{each $step$ in $reason\_path$}
        \IF{$step$ not in $node2id.keys()$}
            \STATE $node2id[step] \gets node\_num$
            \STATE $node\_num \gets node\_num + 1$
        \ENDIF
    \ENDFOR
\ENDFOR

\FOR{each $reason\_path$ in $S_{A_i}$}
    \FOR{each $step$ in $reason\_path$}
        \STATE $start\_node \gets node2id[last\_step]$
        \STATE $end\_node \gets node2id[step]$
        \IF{$(start\_node, end\_node)$ not in $edges$}
            \STATE $edges.add((start\_node, end\_node))$
        \ENDIF
        \STATE $last\_step \gets step$
    \ENDFOR
\ENDFOR

\STATE $G_{A_i} \gets graph(node2id, edges)$
\end{algorithmic}
\end{algorithm}

%% file: Tables/exp.tex
\begin{table*}[ht]
\centering
\begin{tabular}{lcccc}
\hline
                                         & GSM8K         & SVAMP         & ASDiv-a       & \multicolumn{1}{l}{StrategyQA} \\ \hline
Fine-tuning SOTA                         & $57^a$            & $57.4^b$         & $75.3^c$          & $73.9^d$                           \\
9–12 year olds                           & 60            & -             & -             & -                              \\ \hline
{gpt-3.5-turbo:}                     &               &               &               &                                \\
Greedy Decode                            & 72.7          & 78.7          & 93.0          & 65.0                           \\
Self-Consistency (Voting)                & 82.3          & 82.9          & 95.6          & 66.0                           \\
Verifier                                 & 66.9          & 73.1          & 92.8          & 69.3                           \\
Voting Verifier                          & 85.4          & 84.8          & 96.9          & 70.7                           \\
DIVERSE (Step-aware Voting Verifier)     & 85.0          & 85.1          & 96.8          & 66.9                           \\ \hline
\textbf{Reasoning Graph Verifier (Ours)} & \textbf{85.7} & \textbf{85.4} & \textbf{97.0} & \textbf{71.2}                  \\ \hline
\end{tabular}
\caption{The comparison experiment results of \method, other verifiers, and other baselines. We primarily compare \method with other verifiers which are all based on the same generated solutions from \textit{gpt-3.5-turbo}}
\label{tab:exp}
\end{table*}

%% file: Tables/ablation.tex
\begin{table*}[]
\centering
\begin{tabular}{@{}lcccccc@{}}
\toprule
                                         & GSM8K         & \multicolumn{1}{l}{$\bigtriangledown$} & SVAMP         & \multicolumn{1}{l}{$\bigtriangledown$} & ASDiv-a       & \multicolumn{1}{l}{$\bigtriangledown$} \\ \midrule
\textbf{Reasoning Graph Verifier (Ours)} & \textbf{85.7} & -                                      & \textbf{85.4} & -                                      & \textbf{97.0} & -                                      \\ \midrule
w/o solution semantic from base verifier & 81.2          & -4.5                                   & 83.1          & -2.3                                   & 94.3          & -2.7                                   \\
w/o solution scores sum                  & 82.8          & -2.9                                   & 83.2          & -2.2                                   & 95.6          & -1.4                                   \\
w/o reasoning graphs                     & 85.4          & -0.3                                   & 84.8          & -0.6                                   & 96.9          & -0.1                                   \\ \bottomrule
\end{tabular}
\caption{The ablation experiment results of \method. Missing each component leads to a decline in the final result.}
\label{tab:ablation}
\end{table*}

%% file: Tables/llms.tex
\begin{table}[]
\centering
\resizebox{0.48\textwidth}{!}{
\begin{tabular}{lccc}
\hline
                                         & gpt-3.5-turbo & gpt-4         & PaLM-2        \\ \hline
Greedy Decode                            & 72.7          & 87.0          & 53.0          \\
Voting                                   & 82.3          & 94.0          & 71.0          \\
Simple Verifier                          & 66.9          & 89.0          & 36.0          \\
Voting Verifier                          & 85.4          & \textbf{97.0} & 77.0          \\
DIVERSE               & 85.0          & \textbf{97.0} & 75.0          \\ \hline
\textbf{Ours} & \textbf{85.7} & 94.0          & \textbf{78.0} \\ \hline
\end{tabular}
}
\caption{The experimental results of \method with different LLMs.}
\label{tab:llms}
\end{table}

%% file: acl_latex.bbl
\begin{thebibliography}{32}
\providecommand{\natexlab}[1]{#1}

\bibitem[{Agarap(2019)}]{agarap2019deep}
Abien~Fred Agarap. 2019.
\newblock \href {https://arxiv.org/abs/1803.08375} {Deep learning using rectified linear units (relu)}.
\newblock \emph{Preprint}, arXiv:1803.08375.

\bibitem[{Chowdhery et~al.(2022)Chowdhery, Narang, Devlin, Bosma, Mishra, Roberts, Barham, Chung, Sutton, Gehrmann, Schuh, Shi, Tsvyashchenko, Maynez, Rao, Barnes, Tay, Shazeer, Prabhakaran, Reif, Du, Hutchinson, Pope, Bradbury, Austin, Isard, Gur-Ari, Yin, Duke, Levskaya, Ghemawat, Dev, Michalewski, Garcia, Misra, Robinson, Fedus, Zhou, Ippolito, Luan, Lim, Zoph, Spiridonov, Sepassi, Dohan, Agrawal, Omernick, Dai, Pillai, Pellat, Lewkowycz, Moreira, Child, Polozov, Lee, Zhou, Wang, Saeta, Diaz, Firat, Catasta, Wei, Meier-Hellstern, Eck, Dean, Petrov, and Fiedel}]{chowdhery2022palm}
Aakanksha Chowdhery, Sharan Narang, Jacob Devlin, Maarten Bosma, Gaurav Mishra, Adam Roberts, Paul Barham, Hyung~Won Chung, Charles Sutton, Sebastian Gehrmann, Parker Schuh, Kensen Shi, Sasha Tsvyashchenko, Joshua Maynez, Abhishek Rao, Parker Barnes, Yi~Tay, Noam Shazeer, Vinodkumar Prabhakaran, Emily Reif, Nan Du, Ben Hutchinson, Reiner Pope, James Bradbury, Jacob Austin, Michael Isard, Guy Gur-Ari, Pengcheng Yin, Toju Duke, Anselm Levskaya, Sanjay Ghemawat, Sunipa Dev, Henryk Michalewski, Xavier Garcia, Vedant Misra, Kevin Robinson, Liam Fedus, Denny Zhou, Daphne Ippolito, David Luan, Hyeontaek Lim, Barret Zoph, Alexander Spiridonov, Ryan Sepassi, David Dohan, Shivani Agrawal, Mark Omernick, Andrew~M. Dai, Thanumalayan~Sankaranarayana Pillai, Marie Pellat, Aitor Lewkowycz, Erica Moreira, Rewon Child, Oleksandr Polozov, Katherine Lee, Zongwei Zhou, Xuezhi Wang, Brennan Saeta, Mark Diaz, Orhan Firat, Michele Catasta, Jason Wei, Kathy Meier-Hellstern, Douglas Eck, Jeff Dean, Slav Petrov, and Noah Fiedel. 2022.
\newblock \href {https://arxiv.org/abs/2204.02311} {Palm: Scaling language modeling with pathways}.
\newblock \emph{Preprint}, arXiv:2204.02311.

\bibitem[{Cobbe et~al.(2021{\natexlab{a}})Cobbe, Kosaraju, Bavarian, Chen, Jun, Kaiser, Plappert, Tworek, Hilton, Nakano, Hesse, and Schulman}]{cobbe2021gsm8k}
Karl Cobbe, Vineet Kosaraju, Mohammad Bavarian, Mark Chen, Heewoo Jun, Lukasz Kaiser, Matthias Plappert, Jerry Tworek, Jacob Hilton, Reiichiro Nakano, Christopher Hesse, and John Schulman. 2021{\natexlab{a}}.
\newblock Training verifiers to solve math word problems.
\newblock \emph{arXiv preprint arXiv:2110.14168}.

\bibitem[{Cobbe et~al.(2021{\natexlab{b}})Cobbe, Kosaraju, Bavarian, Chen, Jun, Kaiser, Plappert, Tworek, Hilton, Nakano, Hesse, and Schulman}]{cobbe2021training}
Karl Cobbe, Vineet Kosaraju, Mohammad Bavarian, Mark Chen, Heewoo Jun, Lukasz Kaiser, Matthias Plappert, Jerry Tworek, Jacob Hilton, Reiichiro Nakano, Christopher Hesse, and John Schulman. 2021{\natexlab{b}}.
\newblock \href {https://arxiv.org/abs/2110.14168} {Training verifiers to solve math word problems}.
\newblock \emph{Preprint}, arXiv:2110.14168.

\bibitem[{Devlin et~al.(2019)Devlin, Chang, Lee, and Toutanova}]{devlin2019bert}
Jacob Devlin, Ming-Wei Chang, Kenton Lee, and Kristina Toutanova. 2019.
\newblock \href {https://arxiv.org/abs/1810.04805} {Bert: Pre-training of deep bidirectional transformers for language understanding}.
\newblock \emph{Preprint}, arXiv:1810.04805.

\bibitem[{Geva et~al.(2021)Geva, Khashabi, Segal, Khot, Roth, and Berant}]{geva2021strategyqa}
Mor Geva, Daniel Khashabi, Elad Segal, Tushar Khot, Dan Roth, and Jonathan Berant. 2021.
\newblock {Did Aristotle Use a Laptop? A Question Answering Benchmark with Implicit Reasoning Strategies}.
\newblock \emph{Transactions of the Association for Computational Linguistics (TACL)}.

\bibitem[{Google(2023)}]{anil2023palm}
Google. 2023.
\newblock \href {https://arxiv.org/abs/2305.10403} {Palm 2 technical report}.
\newblock \emph{Preprint}, arXiv:2305.10403.

\bibitem[{Huang and Chang(2023)}]{huang2023reasoning}
Jie Huang and Kevin Chen-Chuan Chang. 2023.
\newblock \href {https://arxiv.org/abs/2212.10403} {Towards reasoning in large language models: A survey}.
\newblock \emph{Preprint}, arXiv:2212.10403.

\bibitem[{Kojima et~al.(2023)Kojima, Gu, Reid, Matsuo, and Iwasawa}]{kojima2023large}
Takeshi Kojima, Shixiang~Shane Gu, Machel Reid, Yutaka Matsuo, and Yusuke Iwasawa. 2023.
\newblock \href {https://arxiv.org/abs/2205.11916} {Large language models are zero-shot reasoners}.
\newblock \emph{Preprint}, arXiv:2205.11916.

\bibitem[{Lampinen et~al.(2022{\natexlab{a}})Lampinen, Dasgupta, Chan, Mathewson, Tessler, Creswell, McClelland, Wang, and Hill}]{lampinen-etal-2022-language}
Andrew Lampinen, Ishita Dasgupta, Stephanie Chan, Kory Mathewson, Mh~Tessler, Antonia Creswell, James McClelland, Jane Wang, and Felix Hill. 2022{\natexlab{a}}.
\newblock \href {https://doi.org/10.18653/v1/2022.findings-emnlp.38} {Can language models learn from explanations in context?}
\newblock In \emph{Findings of the Association for Computational Linguistics: EMNLP 2022}, pages 537--563, Abu Dhabi, United Arab Emirates. Association for Computational Linguistics.

\bibitem[{Lampinen et~al.(2022{\natexlab{b}})Lampinen, Dasgupta, Chan, Matthewson, Tessler, Creswell, McClelland, Wang, and Hill}]{lampinen2022can}
Andrew~K Lampinen, Ishita Dasgupta, Stephanie~CY Chan, Kory Matthewson, Michael~Henry Tessler, Antonia Creswell, James~L McClelland, Jane~X Wang, and Felix Hill. 2022{\natexlab{b}}.
\newblock Can language models learn from explanations in context?
\newblock \emph{arXiv preprint arXiv:2204.02329}.

\bibitem[{Li et~al.(2023)Li, Lin, Zhang, Fu, Chen, Lou, and Chen}]{li-etal-2023-making}
Yifei Li, Zeqi Lin, Shizhuo Zhang, Qiang Fu, Bei Chen, Jian-Guang Lou, and Weizhu Chen. 2023.
\newblock \href {https://doi.org/10.18653/v1/2023.acl-long.291} {Making language models better reasoners with step-aware verifier}.
\newblock In \emph{Proceedings of the 61st Annual Meeting of the Association for Computational Linguistics (Volume 1: Long Papers)}, pages 5315--5333, Toronto, Canada. Association for Computational Linguistics.

\bibitem[{Li et~al.(2022)Li, Zhang, Yan, Zhou, Li, Liu, and Cao}]{li-etal-2022-seeking}
Zhongli Li, Wenxuan Zhang, Chao Yan, Qingyu Zhou, Chao Li, Hongzhi Liu, and Yunbo Cao. 2022.
\newblock \href {https://doi.org/10.18653/v1/2022.findings-acl.195} {Seeking patterns, not just memorizing procedures: Contrastive learning for solving math word problems}.
\newblock In \emph{Findings of the Association for Computational Linguistics: ACL 2022}, pages 2486--2496, Dublin, Ireland. Association for Computational Linguistics.

\bibitem[{Loshchilov and Hutter(2019)}]{loshchilov2019decoupled}
Ilya Loshchilov and Frank Hutter. 2019.
\newblock \href {https://arxiv.org/abs/1711.05101} {Decoupled weight decay regularization}.
\newblock \emph{Preprint}, arXiv:1711.05101.

\bibitem[{Miao et~al.(2020)Miao, Liang, and Su}]{miao-etal-2020-diverse}
Shen-yun Miao, Chao-Chun Liang, and Keh-Yih Su. 2020.
\newblock A diverse corpus for evaluating and developing english math word problem solvers.
\newblock In \emph{Proceedings of the 58th Annual Meeting of the Association for Computational Linguistics}, pages 975--984.

\bibitem[{OpenAI(2023)}]{openai2023gpt4}
OpenAI. 2023.
\newblock \href {https://arxiv.org/abs/2303.08774} {Gpt-4 technical report}.
\newblock \emph{Preprint}, arXiv:2303.08774.

\bibitem[{Patel et~al.(2021)Patel, Bhattamishra, and Goyal}]{patel-etal-2021-nlp}
Arkil Patel, Satwik Bhattamishra, and Navin Goyal. 2021.
\newblock \href {https://doi.org/10.18653/v1/2021.naacl-main.168} {Are {NLP} models really able to solve simple math word problems?}
\newblock In \emph{Proceedings of the 2021 Conference of the North American Chapter of the Association for Computational Linguistics: Human Language Technologies}, pages 2080--2094, Online. Association for Computational Linguistics.

\bibitem[{Pi et~al.(2022)Pi, Liu, Chen, Ziyadi, Lin, Fu, Gao, Lou, and Chen}]{pi-etal-2022-reasoning}
Xinyu Pi, Qian Liu, Bei Chen, Morteza Ziyadi, Zeqi Lin, Qiang Fu, Yan Gao, Jian-Guang Lou, and Weizhu Chen. 2022.
\newblock \href {https://doi.org/10.18653/v1/2022.emnlp-main.48} {Reasoning like program executors}.
\newblock In \emph{Proceedings of the 2022 Conference on Empirical Methods in Natural Language Processing}, pages 761--779, Abu Dhabi, United Arab Emirates. Association for Computational Linguistics.

\bibitem[{Rae et~al.(2022)Rae, Borgeaud, Cai, Millican, Hoffmann, Song, Aslanides, Henderson, Ring, Young, Rutherford, Hennigan, Menick, Cassirer, Powell, van~den Driessche, Hendricks, Rauh, Huang, Glaese, Welbl, Dathathri, Huang, Uesato, Mellor, Higgins, Creswell, McAleese, Wu, Elsen, Jayakumar, Buchatskaya, Budden, Sutherland, Simonyan, Paganini, Sifre, Martens, Li, Kuncoro, Nematzadeh, Gribovskaya, Donato, Lazaridou, Mensch, Lespiau, Tsimpoukelli, Grigorev, Fritz, Sottiaux, Pajarskas, Pohlen, Gong, Toyama, de~Masson~d'Autume, Li, Terzi, Mikulik, Babuschkin, Clark, de~Las~Casas, Guy, Jones, Bradbury, Johnson, Hechtman, Weidinger, Gabriel, Isaac, Lockhart, Osindero, Rimell, Dyer, Vinyals, Ayoub, Stanway, Bennett, Hassabis, Kavukcuoglu, and Irving}]{rae2022scaling}
Jack~W. Rae, Sebastian Borgeaud, Trevor Cai, Katie Millican, Jordan Hoffmann, Francis Song, John Aslanides, Sarah Henderson, Roman Ring, Susannah Young, Eliza Rutherford, Tom Hennigan, Jacob Menick, Albin Cassirer, Richard Powell, George van~den Driessche, Lisa~Anne Hendricks, Maribeth Rauh, Po-Sen Huang, Amelia Glaese, Johannes Welbl, Sumanth Dathathri, Saffron Huang, Jonathan Uesato, John Mellor, Irina Higgins, Antonia Creswell, Nat McAleese, Amy Wu, Erich Elsen, Siddhant Jayakumar, Elena Buchatskaya, David Budden, Esme Sutherland, Karen Simonyan, Michela Paganini, Laurent Sifre, Lena Martens, Xiang~Lorraine Li, Adhiguna Kuncoro, Aida Nematzadeh, Elena Gribovskaya, Domenic Donato, Angeliki Lazaridou, Arthur Mensch, Jean-Baptiste Lespiau, Maria Tsimpoukelli, Nikolai Grigorev, Doug Fritz, Thibault Sottiaux, Mantas Pajarskas, Toby Pohlen, Zhitao Gong, Daniel Toyama, Cyprien de~Masson~d'Autume, Yujia Li, Tayfun Terzi, Vladimir Mikulik, Igor Babuschkin, Aidan Clark, Diego de~Las~Casas, Aurelia Guy, Chris Jones,
  James Bradbury, Matthew Johnson, Blake Hechtman, Laura Weidinger, Iason Gabriel, William Isaac, Ed~Lockhart, Simon Osindero, Laura Rimell, Chris Dyer, Oriol Vinyals, Kareem Ayoub, Jeff Stanway, Lorrayne Bennett, Demis Hassabis, Koray Kavukcuoglu, and Geoffrey Irving. 2022.
\newblock \href {https://arxiv.org/abs/2112.11446} {Scaling language models: Methods, analysis \& insights from training gopher}.
\newblock \emph{Preprint}, arXiv:2112.11446.

\bibitem[{Sinha et~al.(2019)Sinha, Sodhani, Dong, Pineau, and Hamilton}]{sinha-etal-2019-clutrr}
Koustuv Sinha, Shagun Sodhani, Jin Dong, Joelle Pineau, and William~L. Hamilton. 2019.
\newblock \href {https://doi.org/10.18653/v1/D19-1458} {{CLUTRR}: A diagnostic benchmark for inductive reasoning from text}.
\newblock In \emph{Proceedings of the 2019 Conference on Empirical Methods in Natural Language Processing and the 9th International Joint Conference on Natural Language Processing (EMNLP-IJCNLP)}, pages 4506--4515, Hong Kong, China. Association for Computational Linguistics.

\bibitem[{Talmor et~al.(2019)Talmor, Herzig, Lourie, and Berant}]{talmor-etal-2019-commonsenseqa}
Alon Talmor, Jonathan Herzig, Nicholas Lourie, and Jonathan Berant. 2019.
\newblock \href {https://doi.org/10.18653/v1/N19-1421} {{C}ommonsense{QA}: A question answering challenge targeting commonsense knowledge}.
\newblock In \emph{Proceedings of the 2019 Conference of the North {A}merican Chapter of the Association for Computational Linguistics: Human Language Technologies, Volume 1 (Long and Short Papers)}, pages 4149--4158, Minneapolis, Minnesota. Association for Computational Linguistics.

\bibitem[{Valmeekam et~al.(2023)Valmeekam, Olmo, Sreedharan, and Kambhampati}]{valmeekam2023large}
Karthik Valmeekam, Alberto Olmo, Sarath Sreedharan, and Subbarao Kambhampati. 2023.
\newblock \href {https://arxiv.org/abs/2206.10498} {Large language models still can't plan (a benchmark for llms on planning and reasoning about change)}.
\newblock \emph{Preprint}, arXiv:2206.10498.

\bibitem[{Wang et~al.(2022)Wang, Zhong, Tang, Wei, Fan, Jiang, Zhou, and Duan}]{wang-etal-2022-logic}
Siyuan Wang, Wanjun Zhong, Duyu Tang, Zhongyu Wei, Zhihao Fan, Daxin Jiang, Ming Zhou, and Nan Duan. 2022.
\newblock \href {https://doi.org/10.18653/v1/2022.findings-acl.127} {Logic-driven context extension and data augmentation for logical reasoning of text}.
\newblock In \emph{Findings of the Association for Computational Linguistics: ACL 2022}, pages 1619--1629, Dublin, Ireland. Association for Computational Linguistics.

\bibitem[{Wang et~al.(2023)Wang, Wei, Schuurmans, Le, Chi, Narang, Chowdhery, and Zhou}]{wang2023selfconsistency}
Xuezhi Wang, Jason Wei, Dale Schuurmans, Quoc Le, Ed~Chi, Sharan Narang, Aakanksha Chowdhery, and Denny Zhou. 2023.
\newblock \href {https://arxiv.org/abs/2203.11171} {Self-consistency improves chain of thought reasoning in language models}.
\newblock \emph{Preprint}, arXiv:2203.11171.

\bibitem[{Wei et~al.(2022)Wei, Wang, Schuurmans, Bosma, Chi, Le, and Zhou}]{wei2022chain}
Jason Wei, Xuezhi Wang, Dale Schuurmans, Maarten Bosma, Ed~Chi, Quoc Le, and Denny Zhou. 2022.
\newblock Chain of thought prompting elicits reasoning in large language models.
\newblock \emph{arXiv preprint arXiv:2201.11903}.

\bibitem[{Xie and Sun(2019)}]{ijcai2019p736}
Zhipeng Xie and Shichao Sun. 2019.
\newblock \href {https://doi.org/10.24963/ijcai.2019/736} {A goal-driven tree-structured neural model for math word problems}.
\newblock In \emph{Proceedings of the Twenty-Eighth International Joint Conference on Artificial Intelligence, {IJCAI-19}}, pages 5299--5305. International Joint Conferences on Artificial Intelligence Organization.

\bibitem[{Xu et~al.(2019)Xu, Hu, Leskovec, and Jegelka}]{xu2018how}
Keyulu Xu, Weihua Hu, Jure Leskovec, and Stefanie Jegelka. 2019.
\newblock \href {https://openreview.net/forum?id=ryGs6iA5Km} {How powerful are graph neural networks?}
\newblock In \emph{International Conference on Learning Representations}.

\bibitem[{Yoran et~al.(2022)Yoran, Talmor, and Berant}]{yoran-etal-2022-turning}
Ori Yoran, Alon Talmor, and Jonathan Berant. 2022.
\newblock \href {https://doi.org/10.18653/v1/2022.acl-long.416} {Turning tables: Generating examples from semi-structured tables for endowing language models with reasoning skills}.
\newblock In \emph{Proceedings of the 60th Annual Meeting of the Association for Computational Linguistics (Volume 1: Long Papers)}, pages 6016--6031, Dublin, Ireland. Association for Computational Linguistics.

\bibitem[{Zhang et~al.(2019)Zhang, Wang, Zhang, Dai, and Shen}]{zhang2019gap}
Dongxiang Zhang, Lei Wang, Luming Zhang, Bing~Tian Dai, and Heng~Tao Shen. 2019.
\newblock \href {https://arxiv.org/abs/1808.07290} {The gap of semantic parsing: A survey on automatic math word problem solvers}.
\newblock \emph{Preprint}, arXiv:1808.07290.

\bibitem[{Zhang et~al.(2020)Zhang, Wang, Lee, Bin, Wang, Shao, and Lim}]{zhang-etal-2020-graph-tree}
Jipeng Zhang, Lei Wang, Roy Ka-Wei Lee, Yi~Bin, Yan Wang, Jie Shao, and Ee-Peng Lim. 2020.
\newblock \href {https://doi.org/10.18653/v1/2020.acl-main.362} {Graph-to-tree learning for solving math word problems}.
\newblock In \emph{Proceedings of the 58th Annual Meeting of the Association for Computational Linguistics}, pages 3928--3937, Online. Association for Computational Linguistics.

\bibitem[{Zhao et~al.(2023)Zhao, Zhou, Li, Tang, Wang, Hou, Min, Zhang, Zhang, Dong, Du, Yang, Chen, Chen, Jiang, Ren, Li, Tang, Liu, Liu, Nie, and Wen}]{zhao2023survey}
Wayne~Xin Zhao, Kun Zhou, Junyi Li, Tianyi Tang, Xiaolei Wang, Yupeng Hou, Yingqian Min, Beichen Zhang, Junjie Zhang, Zican Dong, Yifan Du, Chen Yang, Yushuo Chen, Zhipeng Chen, Jinhao Jiang, Ruiyang Ren, Yifan Li, Xinyu Tang, Zikang Liu, Peiyu Liu, Jian-Yun Nie, and Ji-Rong Wen. 2023.
\newblock \href {https://arxiv.org/abs/2303.18223} {A survey of large language models}.
\newblock \emph{Preprint}, arXiv:2303.18223.

\bibitem[{Zhou et~al.(2023)Zhou, Schärli, Hou, Wei, Scales, Wang, Schuurmans, Cui, Bousquet, Le, and Chi}]{zhou2023leasttomost}
Denny Zhou, Nathanael Schärli, Le~Hou, Jason Wei, Nathan Scales, Xuezhi Wang, Dale Schuurmans, Claire Cui, Olivier Bousquet, Quoc Le, and Ed~Chi. 2023.
\newblock \href {https://arxiv.org/abs/2205.10625} {Least-to-most prompting enables complex reasoning in large language models}.
\newblock \emph{Preprint}, arXiv:2205.10625.

\end{thebibliography}
